# Determining the Characteristic Vocabulary for a Specialized Dictionary using Word2vec and a Directed Crawler

Gregory Grefenstette
Inria Saclay/TAO, Rue Noetzlin - Bât 660
91190 Gif sur Yvette, France
gregory.grefenstette@inria.fr

Lawrence Muchemi
Inria Saclay/TAO, , Rue Noetzlin - Bât 660
91190 Gif sur Yvette, France
lawrence.githiari@inria.fr

**ABSTRACT**

Specialized dictionaries are used to understand concepts in specific domains, especially where those concepts are not part of the general vocabulary, or having meanings that differ from ordinary languages. The first step in creating a specialized dictionary involves detecting the characteristic vocabulary of the domain in question. Classical methods for detecting this vocabulary involve gathering a domain corpus, calculating statistics on the terms found there, and then comparing these statistics to a background or general language corpus. Terms which are found significantly more often in the specialized corpus than in the background corpus are candidates for the characteristic vocabulary of the domain. Here we present two tools, a directed crawler, and a distributional semantics package, that can be used together, circumventing the need of a background corpus. Both tools are available on the web.

## 1. Introduction

Specialized dictionaries (Caruso, 2011) and domain-specific taxonomies are useful for describing the specific way a language is used in a domain, and for general applications such as domain-specific annotation or classification. To create a specialized dictionary, it is first necessary to determine the characteristic vocabulary to be included. These are words that are either specific to the domain, or common words that have specialized usages within the domain. Recent advances using machine learning in natural language processing have led to the development of distributional semantic tools, such as *word2vec,* which use unsupervised training over a large corpus of text to embed words in an *N*-dimensioned vector space (Goldberg and Levy, 2014). These vectors have the desirable property that words that are substitutable, or found in similar contexts, have vectors that are close together in this vector space, and using a distance function, such as cosine distance, reveals words which are semantically similar or related to a given word, or words. To discover the characteristic vocabulary of a domain, it is interesting to see what words are semantically related within that domain. Since the semantic relationships are learned from an underlying corpus, it seems evident that the corpus should be drawn from texts concerning the domain. As a general solution, we have created a directed crawler to build a corpus for any given domain. From this corpus, we can extract the characteristic vocabulary for the domain, and build more complex lexical structures such as taxonomies.

Here, in this article, we present the various pieces that can be assembled to create specialized vocabularies and domain-specific taxonomies. In the next section, we describe how this crawler works. This is followed by a description of one distributional semantics tool, *word2vec*. Then we show how these two tools can be used together to extract the basis of a specialized vocabulary for a domain.

## 2. Building a Directed Crawler

A directed crawler is a web crawler for gathering text corresponding to a certain subject. A web crawler is a program that continuously fetches web pages, starting from a list of seed URLs[1]. Each web page fetched contributes new URLs which are added to the list of the remaining URLs to be crawled. A directed crawler (Chakrabarti et al. 1999) only adds new URLs to this list if the fetched web page passes some filter, such as being written in a given language, or containing certain key words.

In our directed crawler, we begin our crawl using a list of seed URLs from the Open Directory Project[2] (ODP) whose crowd-sourced classification of web pages has been used in many lexical semantic projects (e.g., Osiński and Weiss, 2004; Lee *at al*, 2013; Ševa *et al.*, 2015). To gather the seed list, we send a query concerning the topic of interest, e.g., Fibromyalgia[3], and extract the first 40 URLs returned by the query[4]. These URLs stored in a *ToCrawl* list.

The crawler iterates over this *ToCrawl* list, taking the first URL from the list, fetching the corresponding web page with the Unix *lynx* package[5], and then removing the URL from *ToCrawl*. We do not fetch the same page twice during the crawl, nor more than 100 pages from the same website.

The textual content of the fetched web page is extracted (by the program *delynx.awk,* see release). The page is roughly divided into sentences (*sentencize.awk*), and sentences with at least three English words in a row are retained (*quickEnglish.awk*). Finally, in order to perform the filtering part of the directed crawl, only those pages which contain one or more patterns found in the *Patterns* file are retained. In our released code, the *Patterns* contains upper and lowercase versions of the topic name (e.g. *Fibromyalgia, fibromyalgia*). Retained pages are copied into a *GoodText* directory, and the new URLs found in the retained page (by the *delynx.awk* program) are appended to the *ToCrawl* list. Every time one hundred pages are crawled, the *ToCrawl* list is randomly mixed. The crawl ends when a predefined number of retained pages (e.g., 1000) are found. Collecting 1000 pages for a given topic, using the code delivered, takes around 3 hours on the average.

---

[1] URL stands for *Universal Resource Locator*. URLs most commonly begin with http://… and ftp://…
[2] http://dmoz.org. There are almost 4 million URLs indexed in the ODP catalog, tagged with over 1 million categories. It can be used under the Creative Commons Attribution 3.0 Unported licence
[3] https://www.dmoz.org/search?q=Fibromyalgia
[4] Code found at https://www.lri.fr/~ggrefens/GLOBALEX/
[5] https://en.wikipedia.org/wiki/Lynx_(web_browser)

We have crawled text for 158 autoimmune illnesses[6], and for 266 hobbies[7], in view of creating taxonomies of terms for each topic (Grefenstette, 2015a). Here we will show how to use the distributional semantics tools in *word2vec* to explore these domain-specific corpora, and then show how we build a loose taxonomy automatically.

## 3. Word2vec

Words that appear in similar contexts are semantically related. This is the Distributional Hypothesis (Harris, 1954; Firth 1957). Implementations of this hypothesis have a long history computational linguistics. To find semantically similar nouns using parsed context, Hindle (1990) compared nouns using their frequencies as arguments of verbs as context for comparison, and Ruge (1991) used the frequencies of other words in noun phrases. Frequency of other syntactic relations were used later (Grefenstette, 1994; Lin, 1998), including frequency of appearance in the same lists (Kilgarriff *at al.*, 2004).

In one of the earliest approaches to embedding words in a reduced, fixed-length semantic space, Latent Semantic Indexing (Deerwester *et al.*, 1990) first represented each word by a vector in which each cell value corresponded to the number of times a word appears in a document in some collection. The number of documents in the corpus defined the length of the initial vector. A matrix compression technique, singular value decomposition, allowed them to replace the original word vectors by much shorter, fixed-length vectors (for example, vectors of 100 dimensions). These shorter vectors, or *embeddings* as they are often called now, can be used to recreate the original larger vector with minimal loss of information. As a secondary effect, words whose embeddings are close together, using a cosine measure, for example, to measure the distance, have been found to be semantically similar, as if the singular value matrix reduction mechanism captures some type of "latent semantics."

*Word2vec* (Mikolov *et al.*, 2013) and *GloVe* (Pennington *et al.*, 2014) are two recent tools, among many others (Yin and Schütze, 2015), for creating word embeddings. In *word2vec*, using the *continuous bag of words* setting, word embedding vectors are created by a neural net which tries to guess which word appears in the middle of a context (for example, given the four words preceding and following the word to guess). Using another setting *skip-grams,* the neural net tries to predict the words that appear around a given word. In either case, initial, random word embeddings are gradually altered by the gradient descent mechanism of neural nets, until a stable set is found. Levy and Goldberg (2014) have proved that, with a large number of dimensions in the embedding vectors, and enough iterations, *word2vec* approximates Pointwise Mutual Information (Church and Hanks, 1989; Tunery and Pantel, 2010). *Word2vec* produces "better" results, since it implements other hyperparameters such as generating negative contextual examples, which push unrelated vectors farther apart, and sampling among the positive examples, ignoring some cases, which helps to generalize the vectors since they are not limited to exact contexts (Levy *at al.*, 2015).

Word2vec is memory-efficient and easy-to-use. The code is downloadable from https://code.google.com/p/word2vec/ and it includes scripts for running a number of large scale examples, out of the box. For example, a word2vec script called *demo-word.sh* will download the first 17 million words of Wikipedia and create short embedded vectors for the 71,000 words appearing 5 times or more, in under fifteen minutes on a laptop computer.

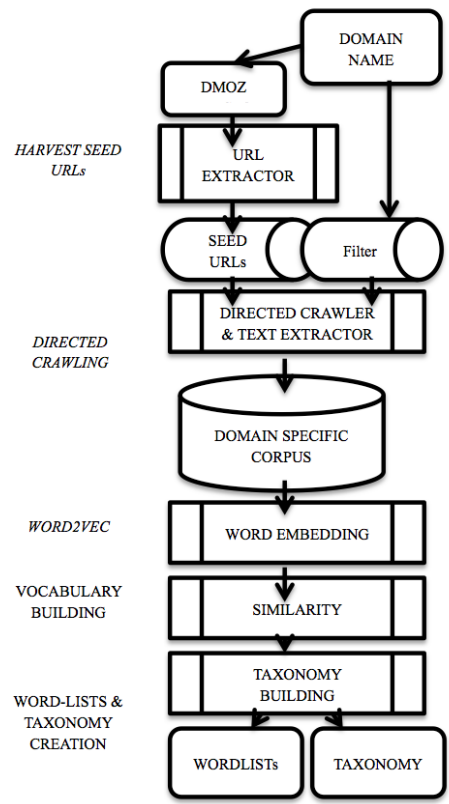

**Figure 1. The structure of our approach, involving a directed crawler to gather text in a given domain, and the use of distributional semantics tool to create the characteristic vocabulary and domain taxonomy.**

## 4. Combining a directed crawl and word2vec

Once a domain specific corpus has been crawled (section 2), *word2vec* can be applied to create fixed size word vectors. The input corpus can be transformed by removing all alphanumeric characters, and transposing uppercase characters to lowercase. This is the case of demo programs delivered in the *word2vec* packages, where, in addition ,all numbers are spelled out as digits (e.g., 19 is written as "one nine") before the word

---

[6] http://www.aarda.org/research-report/ Crawling the 158 topics took about 2 weeks using one computer.
[7] https://en.wikipedia.org/wiki/List_of_hobbies

embedding vectors are trained. Once the vectors are built, one can find the closest words to any word using the *distance* program in the package. For example, using word vectors built from a 750,000 word corpus for fibromyalgia, we find the following words closest to *Fibromyalgia*. The closest the cosine distance is to one, the nearer are the words:

| *Nearest words to Fibromyalgia* | *Cosine distance* |
|---|---|
| pain | 0.573297 |
| symptoms | 0.571838 |
| fatigue | 0.545525 |
| chronic | 0.542895 |
| mysterious | 0.517179 |
| fms | 0.514373 |
| syndrome | 0.514127 |
| cached | 0.508570 |
| treatment | 0.505819 |
| georgia | 0.495497 |
| cfs | 0.492857 |
| overview | 0.492563 |
| referrals | 0.491843 |
| diet | 0.487120 |
| condition | 0.485280 |
| specialists | 0.470644 |
| mcgee | 0.467879 |
| comprehensive | 0.462546 |
| chronicfatigue | 0.462226 |
| fibro | 0.459657 |
| constellation | 0.459147 |
| perplexing | 0.454235 |
| checklist | 0.441451 |
| pinpoint | 0.441292 |
| webmd | 0.441237 |
| controversial | 0.440630 |
| conditions | 0.438186 |
| fm | 0.437467 |

Fibromyalgia is "a rheumatic condition characterized by muscular or musculoskeletal pain with stiffness and localized tenderness at specific points on the body" and many of the words identified by *word2vec* concern its symptoms (*pain, fatigue, , constellation [of symptoms]*) or synonyms (*fibro, fms, chronic-fatigue, fm*) or its characteristics (*mysterious, chronic, perplexing, controversial)* or its treatment (*treatment, referrals, specialists, webmd, diet)*. In order to expand this list, we can find the closest words to each of the 10 most frequent words of length 6 or more*:*

*acceptance, accompanying, aerobic, ailment, amen, anger, anxiety, approach, approaches, appt, arthritic, arthritis-related, biking, bipolar, bloggers, blogspot, brochure, cached, care, cat, cause, causes, celiac, cfs, characterized, cherokeebillie, chronic, clinically, com, common, comprehensive, concurrent, condition, conditioning, conditions, conducted, considerable, constellation, contributing, cortisol, costochondritis, cycles, degenerative, dementia, depressive, dermatomyositis, discomfort, discusses, disease, diseases, disorder, disorders, disturbance, doc, docs, doctors, documentary, dysthymia, ehlers-danlos, elevated, emedicine, emotions, encephalomyelitis, endocrinologist, everydayhealth, excluded, exercises, exercising, exertion, existing, experiencing, expertise, explanations, extent, fatigue, fetus, fibromyalgia, finance, fiona, fischer, flexibility, flu-like, fms, fmsni, focused, frontiers, funding, georgia, guardian, hallmark, hashimoto, hashimotos, healthcare, health-care, homocysteine, hyperthyroidism, hypothyroidism, hypothyroidmom, … , situations, someecards, sought, specialist, sponsors, statistics, stretching, studies, study, subjective, substantial, sufferers, surrounding, swimming, symptomatic, symptoms, syndrome, syndromes, temporary, testosterone, therapy, transforming, treatment, treatments, truths, tsh, underactive, undiagnosed, unrefreshing, valuable, variant, walking, warranty, wealth, websites, wellness, widespread, worsen*

To demonstrate that it is better to use word2vec with a domain specific corpus, rather than a general corpus, consider Tables 1 and 2. In these tables, we compare the closest words found to "pain" and to "examination" in two general corpora, a 10 billion word newspaper corpus, and 17 million word Wikipedia corpus, to 9 domain specific corpora concerning illnesses gathered using the directed crawler of section 2. We see that in the domain specific corpora, the words related to pain are tailored to each illness, whereas the general corpora give words related to pain over a variety of situations. Likewise, for "examination", we can guess from the closest words, what type of medical examinations are used for each illness, whereas the general corpora confuse the academic and judicial senses of "examination" with any medical senses.

### 4.1 Word2vec Trick

Word2vec can also be used to discover the characteristic vocabulary of a domain, given a domain text, such as that crawled by a directed crawler, and a larger, background text not from the same domain. Without modifying the code of *word2vec*, one can make this vocabulary visible using q "trick" of inserting an explicit label inside the domain text only, and learning word vector for this explicit label. Words closest to this explicit, inserted label are those words that are most predictive of the domain label.

Here is how we perform this insertion:
1. Take the domain corpus that has been generated by a directed crawl (as described above in section 2), and remove stopwords[8] and punctuation from the text. Lowercase the resulting text.
2. Insert a new uppercase label between every word in the lowercased domain text.
3. Append the domain text with its explicit label to a large background corpus of English that has been prepared as in step 1
4. Use the resulting text (i.e., domain text with its explicit label and the background text) as input to *word2vec*, and create new word vectors.
5. Use the *distance* program delivered in the *word2vec* package, and find the words that are closest to the uppercase explicit label as the closest words to that domain.

---

[8] http://www.lextek.com/manuals/onix/stopwords2.html, for example

| Google News (10 billion words) | First 17 million words Wikipedia | Domain specific corpora (each about 250k words) | | | | | | | | |
|---|---|---|---|---|---|---|---|---|---|---|
| | | Hypogammaglobulinemia | Vitiligo | Psoriasis | Vasculitis | Uveitis | Neutropenia | Scleroderma | Lupus | Myositis |
| discomfort | neuropathic | nausea | fever | swelling | joint | redness | relief | stiffness | joint | tenderness |
| chronic_pain | nausea | headache | stomach | stiffness | sleeping | tenderness | headache | joint | stiffness | stiffness |
| excruciating_pain | suffering | vomiting | urination | unbearable | stiffness | stiffness | difficulty | physiotherapy | fatigue | aches |
| ache | palpitations | itching | knee | itch | fatigue | ache | legs | aches | tenderness | chills |
| arthritic_pain | headaches | stiffness | vision | stiff | muscle | photophobia | shortness | tiredness | complaints | pains |
| agony | analgesia | flushing | ulcers | joint | aching | ibuprofen | asthenia | relief | aching | malaise |
| soreness | discomfort | chills | decreased | abdominal | tingling | painkillers | epistaxis | appetite | pains | fatigue |
| throbbing_pain | itching | sweats | tooth | weakness | weakness | symptoms | abdominal | shoulder | spasms | redness |
| dull_ache | convulsions | headaches | teeth | joints | myofascial | blurring | chills | swelling | muscle | cramping |
| numbness | ailments | weakness | discolored | vision | shoulders | fatigue | fatigue | mood | swelling | anorexia |
| anxiety | vomiting | dizziness | chest | redness | muscles | spasms | appetite | mobility | fevers | complaint |
| compartmental_syndrome | insomnia | malaise | feeling | botox | diarrhea | pains | weakness | strength | fever | complain |
| burning_sensation | anesthesia | dyspnea | redness | intense | relieve | motion | breath | subacromial | shortness | joint |
| Muscle_spasms | headache | swelling | checker | itching | shortness | sensitivity | edema | exercises | ligaments | aching |
| aches | fibromyalgia | rashes | thickening | headache | appetite | blurred | malaise | tenderness | weakness | swelling |

**Table 1** Words closest to the word "pain", using *word2vec* to generate embedded word vectors from different corpora. The first two columns use word vectors from 100 billion words of newspaper text (Google News), and 17 million words of Wikipedia text, the remaining 9 columns correspond to smaller corpora created by directed crawling. The first two corpora give general, wide-ranging type of pain. The domain specific corpora restrict type of pain to the specified illness.

| Google News (10 billion words) | First 17 million words Wikipedia | Domain specific corpora (each about 250k words) | | | | | | | | |
|---|---|---|---|---|---|---|---|---|---|---|
| | | Hypogammaglobulinemia | Vitiligo | Psoriasis | Vasculitis | Uveitis | Neutropenia | Scleroderma | Lupus | Myositis |
| examinations | examinations | revealed | wood | suspect | exam | slit-lamp | aspirate | exam | laboratory | reveal |
| exam | histological | physical | suspect | determine | physical | biomicroscope | findings | tests | exam | distinguish |
| Examination | baccalaureate | sample | uveitis | diagnosing | piece | reveals | physical | ekg | measurement | careful |
| evaluation | electromyograph | biopsy | physical | determining | histopathological | physical | aspiration | history | evaluation | confirm |
| thorough_examination | autopsy | radiograph | exam | examining | radiological | establishing | investigations | perform | absence | exam |
| exams | study | duodenal | rule | checking | revealed | evaluation | examinations | microscope | microscopic | differentiating |
| inspection | studies | findings | tests | imaging | work-up | revealed | biopsy | changes | biopsy | electrophysiolo |
| dissection | exam | exam | eye | recognize | removal | accomplished | gross | physical | tests | evaluation |
| medico_legal_examin | exams | stool | insufficient | physical | conduct | fundus | exam | confirm | physical | specimen |
| forensic_examination | biopsy | specimen | closed | suspected | specimens | exam | tender | ultrasound | x-ray | radiography |
| assessment | screening | showed | existence | proper | examine | findings | workup | reveal | microscope | scans |
| postmortem | procedure | adenopathies | identifying | confirmation | examined | lamp | careful | sensitive | urinalysis | ultrasound |
| polygraphic_test | tests | mediastinal | perform | dosing | interventional | ophthalmoscopy | specimen | assessed | electrolytes | tomographic |
| examined | accreditation | examinations | trauma | uncertainty | specimen | biomicroscopy | diagnostically | dimensions | ultrasound | histopathology |
| microscopic_examinat | coursework | perform | qualified | biopsy | confirmation | tessler | smear | definitive | repeated | electromyogra |

**Table 2** Words closest to the word "examination", using *word2vec* to generate embedded word vectors from different corpora. The first two columns use word vectors from 100 billion words of newspaper text (Google News), and 17 million words of Wikipedia text, the remaining 9 columns correspond to smaller corpora created by directed crawling. The first two corpora give criminal, newsworthy types of "examination". The domain specific corpora restrict type of pain to the specified illness. Words sorted by nearness to "examination"

## 4.2 Examples

For example, the corpus we crawled for *Vitiligo*[9] comes from 1000 webpages and contains with the following text:

> … Individuals with vitiligo feel self conscious about their appearance and have a poor self image that stems from fear of public rejection and psychosexual concerns ….

After step 1 above, removing stopwords, this text is reduced to

> … individuals vitiligo feel conscious appearance poor image stems fear public rejection psychosexual concerns …

After this step, there were about 150,000 non-stop words in the domain text. In step 2, we insert an explicit label, for example, *VVV*, between each word in the domain text:

> … individuals VVV vitiligo VVV feel VVV conscious VVV appearance VVV poor VVV image VVV stems VVV fear VVV public VVV rejection VVV psychosexual VVV concerns …

In step 3, we append, to this domain text with the explicit labels inserted, another 34 million words coming from a wide-range of English texts with stopwords excluded.

In step 4, this combined text is input into *word2vec* using the default parameters[10] from the demo-word.sh script delivered in the package. This creates word embedding vectors for the 158,000 tokens appearing 5 times or more in the combined text, including our artificially inserted explicit label of step 2.

In step 5, we use the *distance* program of the word2vec package to find the 40 closest words to our artificial label:

> vitiligo, depigmented, repigmenting, leucoderma, bueckert, mequinol, grojean, re-pigmentation, benoquin, repigmentation, depigmentation, bleaching, leukotrichia, lightening, melasma, psoriasis, hair, dpcp, lighten, basc, tacalcitol, complexion, complexions, tanned, camouflage, tattooing, depigmentary, dermablend, de-pigmented, radmanesh, freckle, melanocytes, maquillage, plucking, protopic, eumelanin, alopecia, tans, avrf, leucotrichia

We can expand this list, like in a spreading activation net, by looking for the 5 or 10 closest words to each of these words close to the artificial label. This can raise the number of single-word candidates for the characteristic vocabulary to hundred of words.

## 4.3 From words to phrases

Using the technique described in section 4.1, we gather a certain number of single words that are candidates for the characteristic vocabulary. From our original domain corpus, we can also extract multiword phrases, using a parser such as the Stanford parser, or heuristic methods such as retaining sequences between stopwords. These multiword phrases and their frequencies are filtered through the list of single-word candidates to retain any phrase that contains one of the single word candidates. We stem both the single words and multiword phrases before this filtering is performed. The most frequent multiword phrase thus filtered from our 1000 web page *Vitiligo* corpus include:

> 124 white patch
> 103 vitiligo treatment
> 75 treat vitiligo
> 69 le vitiligo
> 66 vitiligo patient
> 61 skin condit
> 57 autoimmun diseas
> 50 publish onlin
> 48 skin diseas
> 46 segment vitiligo
> 46 gener vitiligo
> 44 white spot

We find about 17,000 multiword and single word candidates for our *Vitiligo* domain in this way.

## 4.4 Filtering candidates

Since we intend to build a taxonomy, we further filter these candidates by only retaining those candidates which appear with another candidate in the domain corpus. This co-occurrence step, reduces the term candidate list to 990.

## 4.5 Building a loose taxonomy

Using the "dog and poodle" intuition, that is, that a term and its hypernym appear often together in the same sentence, and that among co-occurring terms, if one term is much more common (e.g. *dog*) than the other (e.g., *poodle*) then the more common term is the hypernym of the other. We also implement the string inclusion hypothesis, i.e., a subterm of a long term is the hypernym of the longer term. These two strategies were sufficient to place first in the SemEval 2015 Taxonomy task (Grefenstette, 2015b). These two strategies produce a pairwise ordering of retained terms as hypernyms and hyponyms. Here are some examples of these pairings for the Vitiligo domain (words are still stemmed at this point):

> phototherapi>narrowband uvb
> phototherapi>narrowband uvb treatment
> phototherapi>ongo repigment
> phototherapi>parsad
> phototherapi>partner
> phototherapi>perilesion skin
> pigment>caus skin
> pigment>caus whitish patch
> pigment>cell call melanocyt
> pigment cell>melanogenesi

We place the domain term (used to begin the crawl described in section 2) at the root of taxonomy, and place other terms under this root, maintaining order and avoiding loops. We also "unstem" the stemmed terms by producing all variants attested in the original crawled domain corpus. This produces an output such as the following, where the hypernymy relation is expressed using the great-than sign (>):

> vitiligo>basal cell carcinoma>superficial basal cell carcinoma
> vitiligo>bb uvb>targeted bb uvb
> vitiligo>bleaching>skin bleaching

---

[9] Vitiligo is a chronic skin condition characterized by portions of the skin losing their pigment.

[10] -cbow 1 -size 200 -window 8 -negative 25 -hs 0 -sample 1e-4 -threads 20 -binary 1 -iter 15

```
vitiligo>blotches>causes blotches
vitiligo>calcipotriene>calcipotriene ointment
vitiligo>called melanocytes>cells called melanocytes
vitiligo>called melanocytes>white patches appear
vitiligo>camouflage>camouflage creams
vitiligo>camouflage>skin camouflage
vitiligo>camouflage>traditional skin camouflage
vitiligo>causing depigmentation>medical condition causing depigmentation
```

We use these automatically generated taxonomies to annotate user generated text in our personal information system that we are building[11].

## 5. Conclusion

In this paper, we explain how we created a directed crawler that gathers domain-specific text, using open source tools, and also demonstrate how the collected corpus can be exploited by word2vec to discover the basic vocabulary for a given domain.

### 5.1 Acknowledgments

This work is supported by an Advanced Researcher grant from Inria.

---

[11] http://www.slideshare.net/GregoryGrefenstette/pda-2016